\documentclass[prodmode,acmtecs]{acmsmall} 

\usepackage[ruled]{algorithm2e}
\usepackage{tabularx}
\usepackage{graphicx}
\usepackage{graphics}
\usepackage{enumerate,letltxmacro}

\SetAlFnt{\small}
\SetAlCapFnt{\small}
\SetAlCapNameFnt{\small}
\SetAlCapHSkip{0pt}
\IncMargin{-\parindent}

\acmVolume{0}
\acmNumber{0}
\acmArticle{0}
\acmYear{2016}
\acmMonth{0}

\setcopyright{cagov}

\doi{0000001.0000001}

\issn{1234-56789}

\begin{document}

\markboth{Mohammad et al.}{Stance and Sentiment in Tweets}

\title{Stance and Sentiment in Tweets}
\author{Saif M. Mohammad
\affil{National Research Council Canada}
Parinaz Sobhani
\affil{University of Ottawa}
Svetlana Kiritchenko
\affil{National Research Council Canada}
}

\begin{abstract}
We can often detect from a person's utterances whether he/she is in favor of
or against a given target entity---their stance towards the target. 
However, a person may express
the same stance towards a target by using negative or positive language.  
Here for the first time we present a
dataset of tweet--target pairs annotated for both stance and sentiment.
The targets may or may not be referred to in the tweets,
and they may or may not be the target of opinion in the tweets.  
Partitions of this dataset 
were used as training and test sets in a SemEval-2016 shared task competition.
We propose a simple stance detection system that outperforms submissions from all 19 teams that participated in the shared task.
Additionally, access to both stance and sentiment annotations allows us to 
explore several research questions.
We show that while knowing 
the sentiment expressed by a tweet is beneficial for stance classification, it alone is not sufficient.
Finally, we use additional unlabeled data through distant supervision techniques and word embeddings to further improve stance classification.   
\end{abstract}

\terms{Natural Language Processing, Computational Linguistics}


\acmformat{Saif M. Mohammad, Parinaz Sobhani, Svetlana Kiritchenko,
2016. Stance and Sentiment in Tweets}

\begin{bottomstuff}
Author's addresses: Saif M. Mohammad {and} Svetlana Kiritchenko,
National Research Council Canada;
Parinaz Sobhani, 
University of Ottawa.
\end{bottomstuff}

\maketitle

\section{Introduction}
\label{sec:intro}



Stance detection is the task of automatically determining from text whether the author of
the text is in favor of, against, or neutral towards a proposition or target. 
The target may be a person, an organization, a government policy, a movement, a product, etc.
For example, one can infer from Barack Obama's speeches that he is in favor of stricter gun laws in the US.
Similarly, people often express stance towards various target entities through posts on online forums, blogs, Twitter,
Youtube, Instagram, etc.

Automatically detecting stance has widespread applications in information retrieval, text
summarization, and textual entailment. 
Over the last decade, there has been active research in modeling stance. However, most work focuses
on congressional debates \cite{thomas2006get} or debates in online forums
\cite{somasundaran2010recognizing,anand2011cats,walker2012stance,hasan2013stance}. 
Here we explore the task of detecting stance in Twitter---a popular microblogging platform where people often express stance implicitly or explicitly.

%


 The task we explore is formulated as follows: given a tweet text and a target entity (person, organization, issue, etc.),
 automatic natural language systems
 must determine whether the tweeter is in favor of the given
 target, against the given target, or whether neither inference is likely. For example, consider the target--tweet
 pair:
{\small
\begin{quote}
{ Target}: legalization of abortion \hspace{6.7cm} (1)\\
 Tweet: {\it The pregnant are more than walking incubators. They have rights too!}
\end{quote}
 }
\noindent Humans can deduce from the tweet that the tweeter is likely in favor of the target.\footnote{Note that we use `tweet' to refer to the text of the tweet and not to its meta-information.
In our annotation task, we asked respondents to label for stance towards a given target based on the tweet text alone. However, automatic systems may benefit from exploiting tweet meta-information.}

Note that lack of evidence for `favor' or `against', does not imply that the tweeter is neutral towards the target.
It may just mean that we cannot deduce stance from the tweet. In fact, this is a common phenomenon.
On the other hand, the number of tweets from which we can infer neutral stance is expected to be small.
Example:
 {\small
\begin{quote}
{ Target:} Hillary Clinton \hspace{8cm} (2)\\
{ Tweet:} {\it Hillary Clinton has some strengths and some weaknesses.} 
\end{quote}
 }
Stance detection is related to, but different from, sentiment analysis.
Sentiment analysis tasks are formulated as determining whether a piece of text is positive,
negative, or neutral, or determining from text the speaker's opinion and the target of the opinion (the entity towards which opinion is expressed). 
However, in stance detection, systems are to determine favorability
towards a given (pre-chosen) target of interest. The target of interest may not be explicitly mentioned in the text and it may not
be the target of opinion in the text. For
example, consider the target--tweet pair below:
 {\small
\begin{quote}
{ Target:} {\it Donald Trump} \hspace{8.1cm} (3)\\
{ Tweet:}  {\it Jeb Bush is the only sane candidate in this republican lineup.}
\end{quote}
 }
\noindent The target of opinion in the tweet is Jeb Bush, but the given
target of interest is Donald Trump.
Nonetheless,
we can infer that the tweeter is likely to be unfavorable towards Donald Trump.
Also note that, in stance detection, the target can be expressed in different ways which impacts whether the instance
is labeled `favour' or `against'. For example, the target in example 1 could have been phrased as `pro-life movement',
in which case the correct label for that instance is `against'.
Also, the same stance (`favour' or `against') towards a given target can be deduced from
positive tweets and negative tweets.
This interaction between sentiment and stance has not been adequately addressed in past work,
and an important reason for this is the lack of a dataset annotated for both stance and sentiment.

\vspace*{1mm}
\noindent Our contributions are as follows:
\vspace*{-2mm}
\begin{enumerate}
\item  {\it Created a New Stance Dataset}: We created the first dataset of tweets labeled for both stance and sentiment (Section \ref{sec:data} and Section \ref{sec:sent}).
More than 4000 tweets are annotated for whether one can deduce favorable or unfavorable stance towards
one of five targets
`Atheism', `Climate Change is a Real Concern', `Feminist Movement', `Hillary Clinton',
and `Legalization of Abortion'.
Each of these tweets is also annotated for whether the target of opinion expressed in the tweet is the same as the given target of interest. 
Finally, each tweet is annotated for whether it conveys positive, negative, or neutral sentiment.
\vspace*{1mm}
\item {\it Developed an Interactive Visualizer for the Stance Dataset}: We created an online visualizer for our data (Section \ref{sec:stance-ana})
that allows users to explore the data graphically and interactively. 
Clicking on individual components of the visualization, such as a target, stance class, or sentiment class,
filters the visualization to show information pertaining to the selection.
Thus, the visualization can be used to quickly convey key features of the data, for example, 
the percentage of the instances that are labeled as against `Atheism' and yet use positive language,
and also to delve deeper into particular areas of interest of the user.
\vspace*{1mm}
\item {\it Organized a Shared Task Competition on Stance}: Partitions of this stance-annotated data 
were used as training and test sets in the SemEval-2016 shared task competition, Task \#6: Detecting Stance from Tweets \cite{StanceSemEval2016}. 
Participants were provided with 2,914 training instances labeled  for stance for the five targets. The test data included 1,249 instances. All of the stance data is made freely available through the shared task website.
The task received submissions from 19 teams. 
The best performing system obtained an overall average F-score of 67.8. Their approach employed two 
recurrent neural network (RNN) classifiers: the first was trained to predict task-relevant hashtags 
on a large unlabeled Twitter corpus. This network was used to initialize a second RNN classifier, 
which was trained with the provided training data \cite{MITRE}.
\vspace*{1mm}
\item {\it Developed a State-of-the-Art Stance Detection System}: We propose a stance detection system that is much simpler than
the shared task winning system (described above), and yet obtains an even better F-score of 70.3 on
the shared task's test set (Sections 5, 6 and 7).  We use a linear-kernel SVM classifier that relies on features drawn
from the training instances---such as word and character $n$-grams---as well as those obtained using
external resources---such as sentiment features from lexicons and word-embedding features from
additional unlabeled data.  
\vspace*{1mm}
\item {\it Explored Research Questions}: We conduct several experiments to better understand stance detection, and its
interaction with sentiment (Section 6). 
\begin{itemize}
\item We use the gold labels to determine the extent to which stance can be determined simply from sentiment. 
\item We apply the stance detection system (mentioned above in (4)), as a common text classification framework, to determine both stance and sentiment.
We show that while sentiment features are substantially useful for sentiment classification, they are not as effective for stance classification. 
Further, even though both stance and sentiment detection are framed as three-way classification tasks
on a common dataset, 
automatic systems perform markedly
better when detecting sentiment than when detecting stance towards a given target.
\item We show that stance detection is particularly challenging when the tweeter expresses
opinion about an entity other than the target of interest. (The text classification system
performs close to majority baseline for such instances.)
\end{itemize}
\end{enumerate}
\vspace*{-1mm}
\noindent All of the data, an interactive visualization of the data, and the evaluation scripts are available on the task website as well as the homepage for this Stance project.\footnote{http://alt.qcri.org/semeval2016/task6/\\http://www.saifmohammad.com/WebPages/StanceDataset.htm} 




\section{A Dataset for Stance from Tweets}
\label{sec:data}
We now present how we compiled a set of tweets and targets for stance annotation (Section \ref{sec:selection}), and 
the questionnaire and crowdsourcing setup used for stance annotation (Section \ref{sec:quest}). 
An analysis of the stance annotations is presented in Section \ref{sec:stance-ana}.

\subsection{Selecting the Tweet--Target Pairs} 
\label{sec:selection}

Our goal was to create a stance-labeled dataset with the following properties:
{\small
\begin{enumerate}[1:]
\item {The tweet and target are commonly understood by a wide number of people in the US. (The data was 
eventually annotated for stance by respondents living in the US.)}
\item {There must be a significant amount of data for the three classes: `favor', `against', and `neither'.}
\item {Apart from tweets that explicitly mention the target, the dataset should include a significant number of tweets that
express opinion towards the target without referring to it by name.}
 We wanted to include the relatively harder cases for stance detection where the target is referred to in indirect ways
 such as through pronouns, epithets, honorifics, and relationships.
\item Apart from tweets that express opinion towards the target, the dataset should include a significant number of tweets
in which the target of opinion is different from the given target of interest.
 Downstream applications often require stance towards particular pre-chosen targets of interest (for example, a company might be interested in stance towards its product).
 Having data where the target of
 opinion is some other entity (for example, a competitor's product), helps test how well stance detection systems can cope with such instances.
\end{enumerate}
}
\noindent These properties influenced various choices in how our dataset was created.
To help with Property 1, the authors of this paper compiled a list of target entities commonly known in the United States:
 `Atheism', `Climate Change is a Real Concern", `Feminist Movement', `Hillary Clinton', and
 `Legalization of Abortion'. 


 \begin{table}
\tbl{Examples of stance-indicative and stance-ambiguous hashtags that were manually identified.\label{tab:hashtags}}{
 \centering
\resizebox{\textwidth}{!}{
 \begin{tabular}{llll}
 \hline
 \bf Target            	            &  \bf Example              & \bf Example  			&\bf Example \\
 	 		                        &  \bf Favor Hashtag        & \bf Against Hashtag  	&\bf Neutral Hashtag\\ \hline

 Atheism        & \it \#NoMoreReligions   & \it \#Godswill & \it \#atheism \\ 
 Climate Change is Concern & - & \it \#globalwarminghoax & \it \#climatechange \\ 
 Feminist Movement   	&    \it  \it \#INeedFeminismBecaus & \it \#FeminismIsAwful & \it \#Feminism  \\ 
 Hillary Clinton &   \it  \it \#GOHILLARY  &  \it \#WhyIAmNotVotingForHillary & \it \#hillary2016       \\ 
 Legalization of Abortion    	&   \it  \#proChoice & \it \#prayToEndAbortion & \it \#PlannedParenthood \\ \hline
 \end{tabular}
 }
 }
 \vspace*{-2mm}
 \end{table}

We created a small list of hashtags, which we will call {\it query hashtags}, that people use when tweeting about the targets.
We split these hashtags into three categories: (1) {\it favor hashtags}: expected to occur in tweets
expressing favorable stance towards the target (for example, {\it \#Hillary4President}),
 (2) {\it against hashtags:} expected to occur in tweets
expressing opposition to the target (for example, {\it \#HillNo}),
and (3) {\it stance-ambiguous hashtags:} expected to occur in tweets about the target, but are not explicitly indicative of stance 
(for example, {\it \#Hillary2016}).\footnote{A tweet that has a seemingly favorable hashtag towards a target may in fact oppose the target; and this is not uncommon.
 Similarly unfavorable (or against) hashtags may occur in tweets that favor the target.}
 We will refer to favor and against hashtags jointly as {\it stance-indicative (SI) hashtags}.
 Table \ref{tab:hashtags} lists some of the hashtags used for each of the targets.
 (We were not able to find a hashtag that is predominantly used to show favor towards `Climate change is a real concern',
however, the stance-ambiguous hashtags were the source of a large number of tweets eventually labeled `favor' through human annotation.)
Next, we polled the Twitter API to collect over two million tweets containing these query hashtags.
We kept only those tweets where the query hashtags appeared at the end.
We removed the query hashtags from the tweets to exclude obvious cues for the classification task.
Since we only select tweets that have the query hashtag at the end, 
removing them from the tweet often still results in text that is understandable and grammatical.

Note that the presence of a stance-indicative hashtag is not a guarantee that the tweet will have the same stance. 
Further, removal of query hashtags may result in a tweet that no longer expresses the same stance as with the query hashtag.
Thus we manually annotate the tweet--target pairs after the pre-processing described above.
For each target, we sampled an equal number of tweets pertaining to
the favor hashtags, the against hashtags, and the stance-ambiguous hashtags---up to 1000 tweets at most per target. 
This helps in obtaining a sufficient number of tweets pertaining to each of the stance categories (Property 2).
Properties 3 and 4 are addressed to some extent by the fact that removing the query hashtag can sometimes result in tweets
that do not explicitly mention the target. Consider:
{\small
\begin{quote}
Target: Hillary Clinton \hspace{8cm} (4)\\
Tweet: {\it Benghazi must be answered for \#Jeb16}
\end{quote}
 }
\noindent The query hashtags `\#HillNo' was removed from the original tweet, leaving no mention of Hillary Clinton. Yet there
 is sufficient evidence (through references to Benghazi and \#Jeb16) that the tweeter 
 is likely against Hillary Clinton.
Further, conceptual targets such as `Legalization of Abortion' (much more so than person-name targets) have many instances where
the target is not explicitly mentioned.
 For example, tweeters can express stance by referring to {\it foetuses, women's rights, freedoms}, etc.,
 without having to mention {\it legalization} or {\it abortion}.


\subsection{Stance Annotation}
\label{sec:quest}

 The instructions given to annotators for determining stance are shown below.
Descriptions within each option make clear that stance can be expressed in many different ways, for example by explicitly supporting or opposing the target,
  by supporting an entity aligned with or opposed to the target, etc. 
 The second question asks whether the target of opinion in the tweet is the same as the given target of interest.

{\footnotesize
\noindent\makebox[\linewidth]{\rule{\textwidth}{0.4pt}}
Target of Interest: [target entity]\\
Tweet: [tweet with query hashtag removed]\\[-4pt]

\noindent Q1: From reading the tweet, which of the options below is most likely to be true about the tweeter's stance or outlook towards the target:
 \begin{itemize}
 \renewcommand{\labelitemi}{$\bullet$}
\setlength{\itemindent}{3mm}
 \item  We can infer from the tweet that the tweeter supports the target \\[4pt]
  {\it This could be because of any of reasons shown below: 
  \begin{itemize}
   \item[--] the tweet is explicitly in support for the target
   \item[--] the tweet is in support of something/someone aligned with the target, from which we can infer that  the tweeter supports the target
   \item[--] the tweet is against something/someone other than the target, from which we can infer that the tweeter supports the target
  \item[--] the tweet is NOT in support of or against anything, but it has some information, from which we can infer that the tweeter supports the target
  \item[--] we cannot infer the tweeter’s stance toward the target, but the tweet is echoing somebody else's favorable stance towards the target (in a news story, quote, retweet, etc.)
  \end{itemize}
} 
 \vspace*{2mm}
 \item  We can infer from the tweet that the tweeter is against the target \\[4pt]
 {\it This could be because of any of the following:
 \begin{itemize}
   \item[--] the tweet is explicitly against the target
 \item[--] the tweet is against someone/something aligned with the target entity, from which we can infer that the tweeter is against the target
 \item[--] the tweet is in support of someone/something other than the target, from which we can infer that the tweeter is against the target
 \item[--] the tweet is NOT in support of or against anything, but it has some information, from which we can infer that the tweeter is against the target
 \item[--] we cannot infer the tweeter’s stance toward the target, but the tweet is echoing somebody else's negative stance towards the target entity (in a news story, quote, retweet, etc.)
 \end{itemize}
 }
 \vspace*{1mm}
 \item  We can infer from the tweet that the tweeter is neutral towards the target\\ 
 {\it The tweet must provide some information that suggests that the tweeter is neutral towards the target -- the tweet being neither favorable nor against the target is not sufficient reason for choosing this 
 }
\vspace*{1mm}
 \item  There is no clue in the tweet to reveal the stance of the tweeter towards the target (support/against/neutral)\\[-5pt]
\end{itemize}
 
 \vspace*{-1mm}
 \noindent Q2: From reading the tweet, which of the options below is most likely to be true about the focus of opinion/sentiment in the tweet:
\vspace*{-1mm}
 \begin{itemize}
 \renewcommand{\labelitemi}{$\bullet$}
\setlength{\itemindent}{3mm}
 \item  The tweet explicitly expresses opinion about the target
\vspace*{1mm}
 \item  The tweet expresses opinion about something/someone other than the target
\vspace*{1mm}
  \item  The tweet is not expressing opinion about anything
 \end{itemize}
 \vspace*{-4mm}
\noindent\makebox[\linewidth]{\rule{\textwidth}{0.4pt}}
}

\noindent Each of the tweet--target pairs selected for annotation was uploaded on CrowdFlower
for annotation with the questionnaire shown above.\footnote{http://www.crowdflower.com}
We used CrowdFlower's gold annotations scheme for quality control, wherein about 5\% of the data was annotated internally (by the authors). These questions are referred to as gold questions. During crowd annotation, the gold questions are interspersed with other questions, and the annotator is not aware which is which. However, if she gets a gold question wrong, she is immediately notified of it. If the accuracy of the annotations on the gold questions falls below 70\%, the annotator is refused further annotation. This serves as a mechanism to avoid malicious  annotations. 
In addition, the gold questions serve as examples to guide the annotators.

Each question was answered by at least eight respondents.
The respondents gave the task high scores in a post-annotation survey despite noting that
the task itself requires some non-trivial amount of thought: 3.9 out of 5 on ease of task, 4.4 out of 5 on clarity of instructions, and 4.2 out of 5 overall.

 For each target, the data not annotated for stance is used as the {\it domain corpus}---a set of unlabeled tweets
 that can be used to obtain information helpful to determine stance, such as relationships between relevant entities (we explore the use
of the domain corpus in Section \ref{sec:additional-data}).

The number of instances that were marked as neutral stance (option 3 in question 1) was less than 1\%.
Thus, we merged options 3 and 4 into one `neither in favor nor against' option (`neither' for short). 
The inter-annotator agreement was 73.1\% for question 1 (stance) and 66.2\% for Question 2 (target of opinion).\footnote{The overall inter-annotator agreement was calculated by averaging the agreements on all tweets in the dataset. For each tweet, the inter-annotator agreement was calculated as the number of annotators who agree over the majority label divided by the total number of annotators for that tweet.} 
These statistics are for the complete annotated dataset, which includes instances that were genuinely difficult to annotate for stance
 (possibly because the tweets were too ungrammatical or vague) and/or instances that received poor
 annotations from the crowd workers (possibly because the particular annotator did not understand the
 tweet or its context). 
 In order to aggregate stance annotation information from multiple annotators for an instance, rather than opting for simple majority, we marked an instance with a stance only if at least 60\% of the annotators agreed with each other---the instances with less than 60\% agreement were set aside.
\footnote{The 60\% threshold is somewhat arbitrary, but it 
 seemed appropriate in terms of balancing confidence in the majority annotation and having to discard
 too many instances. 
 Annotations for 28\% of the instances did not satisfy this criterion.
 Note that even though we request 8 annotations per questions, some questions may be annotated more than 8 times. Also, a small number of instances received less than 8  annotations.} 
We will refer to this dataset as the {\it Stance Dataset}.
The inter-annotator agreement on this Stance Dataset is 81.85\% for question 1 (stance) and 68.9\% for Question 2 (target of opinion).
The rest of the instances are kept aside for future investigation.

\section{Labeling the Stance Set for Sentiment}
\label{sec:sent}



A key research question this work aims at addressing is the extent to which sentiment is correlated with stance.
To that end, we annotated the same Stance Dataset described above for sentiment in a separate annotation effort
a few months later. 
We followed a procedure for annotation on CrowdFlower similar to that described above for stance,
but now provided only the tweet (no target). 


 Prior work in sentiment annotation has often simply asked the annotator to label a sentence as positive,
 negative, or neutral, largely
leaving the notion when pieces of text should be marked as positive, negative,
  or neutral up to the individual annotators. This is problematic because it can lead
 to differing annotations from annotators for the same text. Further, in several scenarios
 the annotators may be unsure about how best to label the text. Some of these scenarios are listed below. (See \cite{MohammadAnnotation16} for further discussion on the challenges of sentiment annotation.)
\vspace*{-1mm} 
 \begin{itemize}
 \renewcommand{\labelitemi}{$\bullet$}
%
\vspace*{-1mm} 
 \item {\it Sarcasm and ridicule}: Sarcasm and ridicule are tricky from the perspective of assigning
 a single label of sentiment because they can often indicate positive emotional state of the speaker
 (pleasure from mocking someone or something) even though they have a negative attitude
 towards someone or something. An example of ridicule from our dataset:
 {\small
 \begin{quote}
{\it DEAR PROABORTS: Using BAD grammar and FILTHY language and INTIMIDATION 
makes you look ignorant, inept and desperate. \#GodWins}
 \end{quote}
}

\vspace*{1mm} 

\vspace*{1mm} 
 \item {\it Supplications and requests}: Many tweets convey positive supplications to God or positive requests to people
 in the context of a (usually) negative situation. Example from our dataset:
\vspace*{1mm} 
{\small
\begin{quote}
 {\it Pray for the Navy yard. God please keep the casualties minimal. \#1A \#2A \#NRA 
\#COS \#CCOT \#TGDN \#PJNET  \#WAKEUPAMERICA}
 \end{quote}
}

\vspace*{2mm} 
 \item {\it Rhetorical questions}: Rhetorical questions can be treated simply as queries (and thus neutral)
 or as utterances that give away the emotional state of the speaker. For example, consider this example from our dataset of tweets:
 \vspace*{2mm} 
{\small
\begin{quote}
 {\it How soon do you think WWIII \&WWWIV will begin? \#EndRacism}
 \end{quote}
 }
\vspace*{2mm} 
 \noindent On the one hand, this tweet can be treated as a neutral question, but on the other hand, it can be seen as negative because the utterance
 betrays a sense of frustration on the part of the speaker.
 \end{itemize}


\vspace*{-1mm}
\noindent After a few rounds of internal development, we used the questionnaire below to annotate for sentiment:\\[-10pt]

{\footnotesize
\noindent\makebox[\linewidth]{\rule{\textwidth}{0.4pt}}
  \noindent What kind of language is the speaker using?
 \vspace*{-1mm}
  \begin{enumerate}[1.]
\setlength{\itemindent}{3mm}
  \item  the speaker is using positive language, for example, expressions of support, admiration, positive attitude, forgiveness, fostering, success, positive emotional state (happiness, optimism, pride, etc.)
  \item  the speaker is using negative language, for example, expressions of criticism, judgment, negative attitude, questioning validity/competence, failure, negative emotional state (anger, frustration, sadness, anxiety, etc.)
 \item the speaker is using expressions of sarcasm, ridicule, or mockery 
 \item the speaker is using positive language in part and negative language in part
 \item the speaker is neither using positive language nor using negative language
 \end{enumerate}
\vspace*{-4mm}
\noindent\makebox[\linewidth]{\rule{\textwidth}{0.4pt}}
}
\vspace*{-3mm}
 
 \noindent 
The use of the phrases `positive language' and `negative language' encourages respondents to
focus on the language itself as opposed to assigning sentiment based on event outcomes that are beneficial or harmful to the annotator.
 Sarcasm, ridicule, and mockery are included as a separate option (in addition to option 2)
so that respondents do not have to wonder if they should mark such instances as positive or negative.
Instances with different sentiment towards different targets of opinion can be marked with option 4.
Supplications and requests that convey a sense of fostering and support can be marked as positive.
On the other hand, rhetorical questions that betray a sense of frustration and disappointment can be marked as negative.

 Each instance was annotated by at least five annotators on CrowdFlower.
The respondents gave the task high scores in a post-annotation survey: 4.2 out of 5 on ease of task, 4.4 out of 5 on clarity of instructions, and 4.2 out of 5 overall.

For our current work, we chose to
combine options 2 and 3 into one `negative tweets' class
but they can be kept separate in future work if so desired.
We also chose to combine options 4 and 5 into one `neither clearly positive nor clearly negative category' (`neither' for short).
This frames the automatic sentiment prediction task as a three-way classification task, similar to the stance prediction task.
The inter-annotator agreement on the sentiment responses across these three classes was 85.6\%.

\section{Properties of the Stance Dataset}
\label{sec:stance-ana}

We partitioned the Stance Dataset into training and test sets based on
the timestamps of the tweets. All annotated tweets were  ordered
by their timestamps, and the first 70\% of the tweets formed the
training set and the last 30\% formed the test set.
 Table \ref{dist_stance_train_test} shows the number and distribution of instances in the Stance Dataset.
 

\begin{table}
  \tbl{Distribution of instances in the Stance Train and Test sets for Question 1 (Stance).}{
  \centering
\resizebox{\textwidth}{!}{
  \begin{tabular}{lrrrrrrrrr}
    \hline
     & & & \multicolumn{3}{c}{\bf \% of instances in Train} & &  \multicolumn{3}{c}{\bf \% of instances in Test} \\  
\bf Target        & \bf \# total &\bf \# train   &favor &against &neither &\bf \# test &favor &  against &neither  \\ \hline
  Atheism            & 733 & 513    &  17.9     & 59.3     & 22.8 & 220 & 14.5 & 72.7 & 12.7            \\
Climate Change  & 564 & 395 & 53.7 & 3.8 & 42.5 & 169 & 72.8 & 6.5 & 20.7                                  \\ 
Feminist Movement & 949 & 664 &  31.6 & 49.4 & 19.0 & 285 & 20.4 & 64.2 & 15.4                                             \\
Hillary Clinton  & 983 & 689 &  17.1 & 57.0 & 25.8 & 295 & 15.3 & 58.3 & 26.4                                   \\ 
Legal. Abortion    & 933 & 653 & 18.5 & 54.4 & 27.1 & 280 & 16.4 & 67.5 &16.1                         \\ \hline
Total					& 4163 & 2914 & 25.8	& 47.9  & 26.3  & 1249 & 23.1 & 51.8 &25.1 \\ \hline
  \end{tabular}
}
}
  \label{dist_stance_train_test}
\end{table}

\begin{table}
  \tbl{Distribution of instances in the Stance dataset for Question 2 (Target of Opinion).\label{dist_q2}}{
\centering
{\footnotesize
\begin{tabular}{m{7cm}ccc}

\hline
                    & \multicolumn{3}{c}{\bf Opinion towards} \\
 \bf Target   & Target  & Other & No one   \\ \hline
 Atheism                &  49.25    & 46.38  & 4.37                                                             \\
Climate Change is Concern & 60.81                             & 30.50 & 8.69                               \\
Feminist Movement     & 68.28         & 27.40                               &               4.32                                     \\
Hillary Clinton    &  60.32                                         &       35.10        & 4.58                             \\
Legalization of Abortion    & 63.67                                  &  30.97           & 5.36                 \\ \hline
Total       & 61.02 & 33.77 & 5.21 \\ \hline
\end{tabular}
}
}
\end{table}

\begin{table}[t]
\tbl{Distribution of target of opinion across stance labels.\label{dist_q2_stance}}{
\centering
{\footnotesize
\begin{tabular}{m{7cm}ccc}
\hline
                    & \multicolumn{3}{c}{\bf Opinion towards} \\
 \bf Stance  &  Target & Other &  No one   \\ \hline
favor &94.23 & 5.11 & 0.66\\
against   &72.75 & 26.54 & 0.71 \\ \hline
\end{tabular}
}
}
\end{table}

Table \ref{dist_q2} shows the distribution of responses to Question 2 (whether opinion is expressed
directly about the given target). 
Observe that the percentage of `opinion towards other' varies across different targets from 27\% to 46\%.
Table \ref{dist_q2_stance} shows the distribution of instances by target of opinion
for the `favor' and `against' stance labels.
Observe that, as in Example 3, in a number of tweets from which we can infer unfavorable stance towards a target,
the target of opinion is someone/something other than the target (about 26.5\%).
\noindent Manual inspection of the data also revealed that in a number of instances,
the target is not directly mentioned, and yet stance towards the target was determined by the annotators.
About
28\% of the `Hillary Clinton' instances and 67\% of the `Legalization of Abortion' instances
were found to be of this kind---they did not mention `Hillary' or `Clinton' and
did not mention `abortion', `pro-life', and `pro-choice', respectively (case insensitive; with or without hashtag;
with or without hyphen).
Examples (1) and (4) shown earlier are instances of this, and are taken from our dataset.
Some other examples are shown below:
\vspace*{-2mm}
 \begin{quote}
 Target: {Hillary Clinton} \hspace{8.1cm} (5)\\
 Tweet: {\it I think I am going to vote for Monica Lewinsky's Ex-boyfriends Wife}
 
\vspace*{2mm}
 Target: {Legalization of Abortion} \hspace{6.7cm} (6)\\ 
 Tweet: {\it The woman has a voice. Who speaks for the baby? I'm just askin.}
 \end{quote}



 Table \ref{dist_sentiment_train_test} shows the distribution of sentiment labels in the training and test sets. 
Observe that tweets corresponding to all targets, except for `Atheism', are predominantly negative.

\begin{table*}[t!]
  \tbl{Distribution of sentiment in the Stance Train and Test sets.\label{dist_sentiment_train_test}}{
  \centering
  {\footnotesize
  \begin{tabular}{lcccccc}
    \hline
     & \multicolumn{3}{c}{\bf \% of instances in Train} &  \multicolumn{3}{c}{\bf \% of instances in Test} \\  
\bf Target       &    positive & negative & neither &   positive & negative & neither  \\ \hline
  Atheism          &   60.43 & 35.09 & 4.48 & 59.09 & 35.45 & 5.45   \\ 
Climate Change is Concern & 31.65 & 49.62 & 18.73 & 29.59 & 51.48 & 18.93\\ 
Feminist Movement & 17.92 & 77.26 & 4.82 & 19.30 & 76.14 & 4.56 \\ 
Hillary Clinton  & 32.08 & 64.01 & 3.92 & 25.76 & 70.17 & 4.07 \\ 
Legalization of Abortion    & 28.79 & 66.16 & 5.05 & 20.36 & 72.14 & 7.5\\ \hline
Total                  & 33.05 & 60.47 & 6.49 &  29.46 & 63.33 & 7.20\\ \hline
  \end{tabular}
}
  }
\end{table*}


\begin{figure}
\centerline{\includegraphics[width=\textwidth]{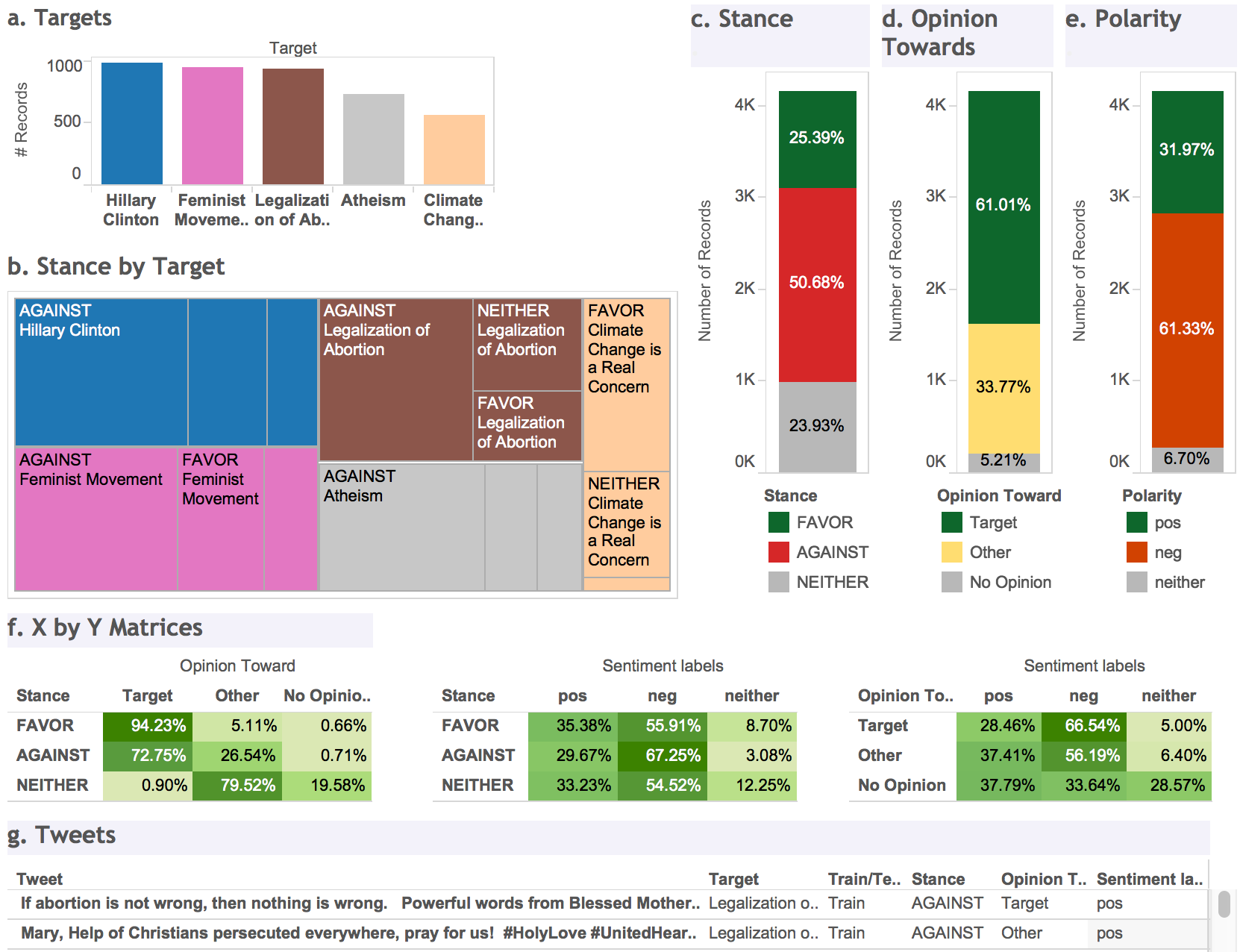}}
\caption{Screenshot of an Interactive Visualization of the Stance Dataset.}
\label{fig:viz}
\end{figure}

To allow ease of exploration of the Stance Dataset we created an interactive
visualization using Tableau---a software product that provides a graphical interface, menu options, and drag-and-drop mechanisms to upload databases and create sophisticated visualizations.\footnote{http://www.tableau.com} 
Figure \ref{fig:viz} shows a screenshot of the visualization of the Stance Dataset. It has several components (a. through g.) that
we will describe below. On the top left, component a., is a bar graph showing the number of instances pertaining to each of the targets in the dataset.
The visualization component b. (below a.), known as a treemap, shows tiles corresponding to each target--stance combination.
The size (area) of a tile is proportional to the number of instances corresponding to that target--stance combination.
This component quickly shows that for most of the targets, the Stance Dataset has more data for `against' than `favor' and `neither'.
The three stacked bars on the top right (c., d., and e.) show the proportion of instances pertaining to the classes of stance, opinion target,
and polarity, respectively.
Observe that they convey to the viewer that a majority of the instances are labeled as `against' the targets of interest,
expressing opinion towards the target of interest, and having negative polarity.

The `f. X by Y Matrices' component of the visualization shows three matrices pertaining to:
stance classes and opinion towards classes, stance classes and polarity classes, and opinion towards classes and polarity classes.
The cells in each of these matrices show the percentage of instances with labels corresponding to that cell (the percentages across 
each of the rows sums up to 100\%.)
Examination of this matrix reveals that favorable stance
is usually expressed by expressing opinion directly about the target (94.23\%), but that percentage is markedly smaller
for instances that are labeled `against the target' (72.75\%).
The visualization component g.\@ at the bottom shows all of the tweets, targets, and manual annotations.

All of the visualization components allow filtering of data by clicking on areas of interest.
For example, clicking on the `Hillary Clinton'
bar updates all other visualization components to show information pertaining to only those
instances that have `Hillary Clinton' as target. 
Clicking on multiple items results in an `AND'ing of the selected filter criteria. For example,
clicking on the target `Atheism', stance `against', and polarity `positive' will show information pertaining to instances
that have Atheism as target, `against' the target stance, and positive polarity labels.

The `Tweets' component at the bottom also filters out information so that one can see examples pertaining to their selection.
Some of the items in individual visualizations may not have enough space to have visible labels (for example,
the Hillary Clinton--Favor tile in the `Stance by Target' treemap). However, hovering over any item with the mouse
reveals the label in a hover box.
We hope that the visualization will help users easily explore aspects of the data they are interested in.

\vspace*{-1mm}
\section{A Common Text Classification Framework for Stance and Sentiment}
\label{sec:comm}

Past work has shown that
the most useful features for sentiment analysis are word and character $n$-grams and sentiment lexicons,
whereas others such as negation features, part-of-speech features, and punctuation have a smaller impact
\cite{SemEval2013task2,MohammadSemEval2013,Kiritchenko2014,rosenthal2015semeval}.
These features may be useful in stance classification as well; however, it is unclear which features will be more useful (and to what extent). 
Since we now have a dataset annotated for both stance and sentiment,
we create a common text classification system (common machine learning framework and common features)
and apply it to the Stance Dataset for both stance and sentiment classification. 

There is one exception to the common machine learning framework.
The words and concepts used in tweets corresponding to the three stance categories are not expected to 
generalize
across the targets. Thus, the stance system learns a separate model from training data pertaining to each of the targets.\footnote{Experiments 
with a stance system that learns a single model from all training tweets showed lower results.}
Positive and negative language tend to have sufficient amount of commonality regardless of topic of discussion, and hence
sentiment analysis systems traditionally learn a single model from all of the training data \cite{liu2015sentiment,Kiritchenko2014,rosenthal2015semeval}.
Thus, our sentiment experiments are also based on a single model trained on all of the Stance Training set.

Tweets are tokenized and part-of-speech tagged with the
CMU Twitter NLP tool \cite{Gimpel11}.
We train a linear-kernel Support Vector Machine (SVM) classifier on the Stance Training set.
SVMs 
have proven to be effective on text categorization tasks and
robust on large feature spaces. We use the SVM implementation provided by the scikit-learn Machine Learning
library \cite{scikit-learn}.



\noindent The features used in our text classification system are shown below:
\vspace*{-3mm}
\begin{itemize}
\renewcommand{\labelitemi}{$\bullet$}
\item {\it $n$-grams}:  presence or absence of contiguous sequences of 1, 2 and 3 tokens (word $n$-grams); presence or absence of contiguous sequences of 2, 3, 4, and 5 characters (character $n$-grams);
\item {\it sentiment (sent.)}: The sentiment lexicon features are derived from three 
manually created lexicons: NRC Emotion Lexicon \cite{mohammad2010emotions}, Hu and Liu 
Lexicon \cite{hu2004mining}, and MPQA Subjectivity Lexicon \cite{Wilson05}, and two
automatically created, tweet-specific, lexicons: NRC Hashtag Sentiment and NRC Emoticon (a.k.a. Sentiment140)
\cite{Kiritchenko2014};
\item {\it target}: presence/absence of the target of interest in the tweet;\footnote{For instance, for `Hillary Clinton' the mention of either `Hillary' or `Clinton' (case insensitive; with or without hashtag) in the tweet shows the presence of target.}
\item {\it POS}: the number of occurrences of each part-of-speech tag (POS); 
\item {\it encodings (enc.)}: 
presence/absence of positive and negative emoticons, hashtags, characters in upper case, elongated words (e.g., {\it sweeettt}), and punctuations
such as exclamation and question marks. 
\end{itemize}
\vspace*{-2mm}
\noindent The SVM parameters are tuned using 5-fold cross-validation on
Stance Training set.
We evaluate the learned models on the 
Stance Test set.  
As the evaluation measure, we use the average of the F1-scores (the harmonic mean of precision and recall) for the two main classes:\footnote{A similar metric was used in the past for sentiment analysis---SemEval 2013 Task 2 \cite{SemEval2013task2}.}\\
\hspace*{6mm} For stance classification:
\vspace*{-3mm}
\begin{equation}
F_{\it average}= \frac{F_{\it favor}+F_{\it against}}{2}
\end{equation}

\hspace*{3mm} For sentiment classification:
\vspace*{-3mm}
\begin{equation}
F_{\it average}= \frac{F_{\it positive}+F_{\it negative}}{2}
\end{equation}

\noindent Note that 
$F_{average}$ can be determined for all of the test instances or for each target data separately. We will refer to the $F_{average}$ obtained through the former method as {\it F-micro-across-targets} or {\it F-microT} (for short).
On the other hand, the $F_{average}$ obtained through the latter method, that is, by averaging the $F_{average}$ calculated for each target separately, will be called {\it F-macro-across-targets} or {\it F-macroT} (for short).
Systems that perform relatively better on the more frequent target classes will obtain higher F-microT scores. On the other hand, to obtain a high F-macroT score a system has to  perform well on all target classes.

 Note that this measure does not give any credit for correctly classifying `neither' instances.
 Nevertheless, the system has to correctly predict all three classes 
 to avoid being
 penalized for misclassifying `neither' instances as `favor' or `against'.

\renewcommand{\tabcolsep}{0.1cm}

\begin{table*}[t!]
\tbl{Stance Classification: F-scores obtained for each of the targets (the columns) by the benchmark systems and our classifier. Macro- and micro-averages across targets  are also shown. Highest scores 
are shown in bold.\label{tab:stance-results}}{
\centering
\resizebox{\textwidth}{!}{
\begin{tabular}{lrrrrrrrrr}
\hline
 	&Atheism	& Climate 	& Feminist  	& Hillary   	&Legal. of	&F- &F-\\
\bf Classifier  	&    	& Change 	& Movemt. 	& Clinton	&Abortion  	&macroT &microT  \\ \hline
{\it I. Benchmarks}	&\\
   $\;\;\;$ a. Random  & 31.1 & 27.8 & 29.1 & 33.5 & 31.1 &30.5 & 33.3   \\ 
 $\;\;\;$ b. Majority   	& 42.1 	& 42.1 	& 39.1 	& 36.8 	& 40.3 	& 40.1   & 65.2\\
 $\;\;\;$ c. First in shared task & 61.4 & 41.6 & 62.1 & 57.7 & 57.3 &56.0 &67.8 \\
 $\;\;\;$ d. Oracle Sentiment & 65.8 & 34.3 & 61.7 & 62.2 & 41.3 &53.1 & 57.2 \\
  $\;\;\;$ e. Oracle Sentiment and Target  & 66.2 & 36.2 & 63.7 & 72.5 & 41.8 & 56.1 & 59.6\\
 \it II. Our SVM classifier	&\\	
 $\;\;\;$   a. $n$-grams	&     65.2 	&\bf 42.4 	& 57.5 	& 58.6 	&\bf 66.4 	& 58.0  & 69.0   \\		
  $\;\;\;$   b. a. + POS	& \bf    65.8 	& 41.8 	& \bf 58.7 	& 57.6 	& 62.6 	& 57.3  & 68.3   \\			
  $\;\;\;$   c. a. + encodings 	& 65.7 &  42.1	& 57.6 	& 58.4 	& 64.5 	& 57.6	 & 68.6\\
 $\;\;\;$   d. a. + target 	& 65.2 &  42.2	& 57.7 	& 60.2 	& 66.1 	&\bf 58.3	  & \bf 69.1\\
 $\;\;\;$ e. a. + sentiment 	& 65.2 	& 40.1	& 54.5 	& \bf 60.6 	& 61.7 	& 56.4  & 66.8\\
\hline
\end{tabular}
}
}
\end{table*}


\section{Results Obtained by Automatic Systems}
\label{sec:results}
We now present results obtained by the classifiers described above
on detecting stance and sentiment on the Stance Test set.
In this section, we focus on systems that use only the provided training data and existing resources such as sentiment lexicons. In Section 7, we conduct experiments with systems that use additional unlabeled (or pseudo-labeled) tweets as well. 

\subsection{Results for Stance Classification}

We conducted 5-fold cross-validation on the stance training set to determine usefulness of each of the features discussed above. Experiments on the test set showed the same patterns. Due to space constraints, we show results only on the test set --- Table \ref{tab:stance-results}. 
Rows I.a. to I.e.\@ present benchmarks.
Row I.a.\@ shows results obtained by a random classifier 
(a classifier that randomly assigns a stance class to each instance), and Row I.b.\@ shows results obtained by the majority classifier
(a classifier that simply labels every instance with the majority class).\footnote{Since our evaluation measure is the average of the F1-scores for the `favor' and `against' classes, the random benchmark depends on the distribution of these classes and is different for different targets. The majority class is determined separately for each target.} 
Observe that the F-microT for the majority classifier is rather high. 
This is mostly due to the differences in the class distributions for the five targets: for most of the targets the majority of the instances are labeled as `against' whereas for target `Climate Change is a Real Concern' most of the data are labeled as `favor'. Therefore, the F-scores for the classes `favor' and `against' are more balanced over all targets than for just one target. 
Row I.c.\@ shows results obtained by the winning system (among nineteen participating teams) in the 2016 SemEval shared task on this data (Task \#6).\\[-7pt]  

\noindent {\bf Results of Oracle Sentiment Benchmarks:}\\
The Stance Dataset with labels for both stance and sentiment allows us, for the first time,
to conduct an experiment to determine the extent to which stance detection can be solved
with sentiment analysis alone. Specifically, we determine the performance of an oracle
system that assigns stance as follows:
For each target, select a sentiment-to-stance assignment 
(mapping all positive instances to `favor' and all negative instances to `against'
OR mapping all positive instances to `against' and all negative instances to `favor') 
that maximizes the F-macroT score.\footnote{Tweets with sentiment label `neither' are always mapped to the stance label `neither'.}
We call this benchmark the Oracle Sentiment Benchmark. 
This benchmark is informative because it gives an upper bound of the F-score one can expect when using a traditional sentiment analysis
system for stance detection by simply mapping sentiment labels to stance labels.\footnote{This is an upper bound because gold sentiment labels are used
and because the sentiment-to-stance assignment is made in a way that is not usually available in real-world scenarios.}

In our second sentiment benchmark, Oracle Sentiment and Target, we include the information on the target of opinion. 
Recall that the Stance Dataset is also annotated for whether the target of opinion is the same as the target of interest. 
We use these annotations in the following way: If the target of opinion is the same as the target of interest, the stance label is assigned in the same way as in the Oracle Sentiment Benchmark; if the target of opinion is some other entity (whose relation to the target of interest we do not know), we select the sentiment-to-stance assignment from the three options: mapping all positive instances to `favor' and all negative instances to `against' OR mapping all positive instances to `against' and all negative instances to `favor' OR mapping all instances to 'neither'; tweets with no opinion are assigned the 'neither' class. 
Again, the selection is done as to optimize the F-macroT score. 
This benchmark indicates the level of performance one can expect when a sentiment analysis system is supplemented with the information on the target of opinion. 

Rows I.d.\@ and I.e.\@ in Table \ref{tab:stance-results} show the F-scores obtained by the Oracle Sentiment Benchmarks on the Stance Test set.
Observe that the scores are higher than the majority baseline for most of the targets, but yet much lower than 100\%. This shows that even though
sentiment can play a key role in detecting stance, sentiment alone is not sufficient.\\[-7pt]

\noindent {\bf Results Obtained by Our Classifier:}\\
Row II.a.\@ shows results obtained by our classifier with $n$-gram features alone.
Note that not only are these results markedly higher than the majority baseline, most of these results are also higher 
than the best results obtained in SemEval-2016 Task 6 (I.c.) and the Oracle benchmarks (I.d. and I.e.). 
Surpassing the best SemEval-2016 results with a simple SVM-ngrams implementation 
is a little surprising, but it is possible that the SemEval teams did not implement a strong $n$-gram baseline such as that presented here, or obtained better results using additional features in cross-validation that did not translate to better results when applied to the test set. (The best systems in SemEval-2016 Task 6 used recurrent neural networks and word embeddings.) 

Rows II.b. through II.e. show results obtained when using other features (one at a time) over and above the $n$-gram features.
Observe that adding the target features leads to small improvements, but adding all other features (including those drawn from sentiment lexicons) does not improve results. 
Additional combinations of features such as `$n$-grams + target + sentiment' also did not improve the performance (the results are not shown here).

\renewcommand{\tabcolsep}{0.2cm}
\begin{table}
\tbl{Stance Classification: F-scores obtained on tweets with opinion towards the target and on tweets with opinion towards another entity.\label{tab:stance-target}}{
\centering
{\footnotesize
\begin{tabular}{lrrcrr}

\hline
				& \multicolumn{2}{c}{\bf F-macroT} & & \multicolumn{2}{c}{\bf F-microT} \\
 \bf Classifier  &To Target &To Other & &To Target &To Other\\ \hline

{\it Benchmarks}	&\\
   $\;\;\;$ a. Random  & 34.3 & 20.0  & & 37.4 & 21.6\\ 
 $\;\;\;$ b. Majority   	& 44.1 	& 28.6 &	& 71.2	& 41.3 	 \\
 $\;\;\;$ c. First in shared task & 59.7 	& 35.4	& 	& 72.5 	& 44.5 \\
 $\;\;\;$ d. Oracle Sentiment &  61.0	& 30.0	& 	& 65.3 	& 33.3 \\
  $\;\;\;$ e. Oracle Sentiment and Target  &  61.0	& 15.7	& & 65.3	& 28.8 	 \\
 \it Our SVM classifier	&\\	
 $\;\;\;$   a. $n$-grams + target	&  62.5	& 37.9	& 	& 75.0 	& 43.0 \\
\hline
\end{tabular}
}
}
\vspace*{-3mm}
\end{table}

Table \ref{tab:stance-target} shows stance detection F-scores obtained by our classifier (SVM with $n$-gram and target features) over the subset of tweets that express opinion towards the given target and the subset of tweets that express opinion towards another entity.\footnote{The results for the Oracle Sentiment and Target benchmark are low on the subset of tweets that express opinion towards another entity since for some of the targets all instances in this subset are assigned to the 'neither' class, and therefore the F-score for such targets is zero on this subset.} 
Observe that the performance of the classifier is
considerably better for tweets where opinion is expressed towards the target, than otherwise.
Detecting stance towards a given target from tweets that express opinion about some other entity
has not been addressed sufficiently in our research community, and we hope our dataset will encourage more work to address this challenging task. 

\renewcommand{\tabcolsep}{0.1cm}

\begin{table*}[t!]
\tbl{Sentiment Classification: F-scores obtained for each of the targets (the columns) by the benchmark systems and our classifier. Macro- and micro-averages across targets  are also shown. Highest scores 
are shown in bold. Note 1: `enc.' is short for encodings; `sent.' is short for sentiment. Note 2: Even though results are shown for subsets of the test set corresponding to targets, unlike stance classification, for sentiment, we do not train a separate model for each target.\label{tab:sent-results}}{
\centering
{\footnotesize
\begin{tabular}{lrrrrrrrrr}
\hline
 	&Atheism	& Climate 	& Feminist  	& Hillary   	&Legal. of	&F- &F-\\
\bf Classifier  	&    	& Change 	& Movemt. 	& Clinton	&Abortion  	&macroT &microT  \\ \hline
{\it I. Benchmarks}	&\\
   $\;\;\;$ a. Random  & 33.8 & 29.6 & 37.3 & 32.1 & 41.1 &34.8 & 35.7   \\ 
 $\;\;\;$ b. Majority   	& 26.2 	& 34.0 	& 43.2 	& 41.2 	& 41.9 	& 37.3   & 38.8\\
 \it II. Our SVM classifier	&\\	
 $\;\;\;$   a. $n$-grams	&     69.7 	& 66.9 	& 65.3 	& 75.9 	& 73.2 	& 70.2  & 73.3   \\		
  $\;\;\;$   b. a. + POS	& 73.3 & 64.2 & 69.9 & 75.1 & 74.5 & 71.4 & 74.4  \\			
  $\;\;\;$   c. a. + encodings 	& 69.8 & 66.2 & 67.8 & 75.9 & 72.9 & 70.5 & 73.5\\
 $\;\;\;$ d. a. + sentiment 	&\bf 76.9 	&\bf 72.6	&\bf 70.9 	& 80.1 	&\bf 80.7 	&\bf 76.4  & \bf 78.9\\
$\;\;\;$ e. b. + enc. + sent. & 76.3 & 70.6 & 70.5 & \bf 80.7 & 79.2 & 75.5 & 78.1\\
\hline
\end{tabular}
}
}
\end{table*}

\renewcommand{\tabcolsep}{0.2cm}
\begin{table}
\tbl{Sentiment Classification: F-scores obtained on tweets with opinion towards the target and on tweets with opinion towards another entity.\label{tab:target-sent}}{
\centering
{\footnotesize
\begin{tabular}{lrrcrr}

\hline
				& \multicolumn{2}{c}{\bf F-macroT} & & \multicolumn{2}{c}{\bf F-microT} \\
 \bf Classifier  &To Target &To Other & &To Target &To Other\\ \hline

{\it I. Benchmarks}	&\\
   $\;\;\;$ a. Random  & 33.8 & 36.6  & & 29.2 & 34.6\\ 
 $\;\;\;$ b. Majority   	& 38.4 	& 36.1 &	& 40.0	& 36.9 	 \\
 \it II. Our SVM classifier	&\\	
 $\;\;\;$   a. $n$-grams + sentiment	&  75.8	& 76.2	& 	& 78.9 	& 79.0 \\
\hline
\end{tabular}
}
}
\end{table}



 
 



\subsection{Results for Sentiment Classification}


Table \ref{tab:sent-results} shows F-scores obtained by 
various automatic systems on the sentiment labels of the Stance Test set.
Observe that 
the text classification system obtains markedly higher scores on sentiment prediction than on predicting stance. 

Once again a classifier trained with $n$-gram features alone obtains results markedly higher than the baselines (II.a.). However, here (unlike as in the stance task) sentiment lexicon features provide marked further improvements (II.d). Adding POS and encoding features over and above $n$-grams results in modest gains (II.b. and II.c.) 
Yet, a classifier trained with all features (II.e.) does not outperform the classifier trained with only $n$-gram and sentiment features (II.d.).


Table \ref{tab:target-sent} shows the performance of
the sentiment classifier (SVM with $n$-grams and sentiment features) on tweets that express opinion towards the given target and those that express opinion about another entity. 
Observe that the sentiment prediction performance (unlike stance prediction performance) is similar on the two sets of tweets.
This shows that the two sets of tweets are not qualitatively different in how they express opinion. However, since one
set expresses opinion about an entity other than the target of interest, detecting stance towards the target of interest from them is 
notably more challenging.

\section{Stance Classification using Additional Unlabeled Tweets}
\label{sec:additional-data}
Classification results can usually be improved by using more data in addition to the training set.
In the sub-sections below we explore two such approaches when used for stance classification: distant supervision and word embeddings.

\subsection{Distant Supervision}
\label{sec:distsup}
{\it Distant supervision} is a method of supervised text classification wherein the training data
is automatically generated using certain indicators present in the text.
For example, \citeN{Go2009} extracted tweets that ended with emoticons `:)' and `:('.
Next, the emoticons were removed from the tweets and the remaining portions of the tweets
were labeled positive or negative depending on whether they originally had `:)' or `:(', respectively.
Central to the accuracy of these sentiment labels is the idea that emoticons are often redundant
to the information already present in the tweet, that is, for example, a tweet that ends with a  `:)' emoticon
likely conveys positive sentiment even without the emoticon. \citeN{Mohammad12} and \citeN{kunneman2014predictability} tested 
a similar hypothesis  for emotions conveyed by hashtags at the end of a tweet and the rest of the tweet.
In Section \ref{sec:red}, we test the validity of the hypothesis that in terms of conveying stance,
stance-indicative hashtags are often redundant to the information already present in the rest of the tweet.
In Section \ref{sec:dsupexp}, we show how we compiled additional training data using stance-indicative hashtags, and used it for stance classification.

\subsubsection{Redundancy of Stance-Indicative Hashtags}
\label{sec:red}

Given a target, stance-indicative (SI) hashtags can be determined manually (as we did to collect tweets).
We will refer to the set we compiled as {\it Manual SI Hashtags}.
Note that this set does not include the manually selected stance-ambiguous hashtags. 
Also, recall that the Manual SI Hashtags were removed from tweets prior to stance annotation. 

To determine the extent to which an SI hashtag is 
redundant to the information already present in the tweet (in terms of conveying stance), we created a stance classification system
that given a tweet-target instance from the Stance Test set, assigns to it the stance associated with the
hashtag it originally had. Table \ref{tab:red-manl} shows the accuracy 
of Favor--Against Classification on the 555 instances of the Stance Test set
which originally had the manually selected SI hashtags.
Observe that the accuracy is well above the random baseline 
indicating that many SI hashtags are used redundantly in tweets (in terms of conveying stance). This means that these hashtags can be used to automatically collect additional, somewhat noisy, stance-labeled training data.

%

\begin{table*}
\tbl{Accuracy of Favor--Against Classification on the 555 instances of the Stance Test set 
which originally had the manually selected stance-indicative hashtags.\label{tab:red-manl}}{ 
\centering
{\footnotesize
\begin{tabular}{m{9.1cm}rrr}
\hline
 \bf System  					& Accuracy  \\ \hline
   a. Random Baseline			& 50.0\\
   b. Hashtag-based classification 	& 68.3\\
                   \hline
\end{tabular}
}
}
\end{table*}


\begin{table}
\tbl{Examples of SI hashtags compiled automatically from the Stance Training set.\label{Tab:hashtag_example}}{
\centering
{\footnotesize
\begin{tabular}{lrll}
\hline
 \bf Target 			&  \# hashtags 			&  Favor hashtag 		&  Against hashtag  \\ \hline
Atheism                 & 14                        & \#freethinker  			& \#prayer        \\
Climate Change          & 9                         & \#environment  			& -       \\
Feminist Movement                & 10                 		& \#HeForShe      			& \#WomenAgainstFeminism             \\
Hillary Clinton         & 19                   		& -   						& \#Benghazi \\
Legal. Abortion                & 18                        & \#WomensRights 			& \#AllLivesMatter   \\ \hline
\end{tabular}
}
}
\end{table}

\subsubsection{Classification Experiments with Distant Supervision}
\label{sec:dsupexp}

If one has access to tweets labeled with stance, then one can estimate how well a hashtag can predict stance using the following score:
\begin{equation}
H ({\it hashtag}) = max_{stance\_label \in \{favor,against\}} \frac{{\it freq} ({\it hashtag}, {\it stance\_label})}{{\it freq} ({\it hashtag})}
\end{equation}
where ${{\it freq}({\it hashtag})}$ is the number of tweets that have that particular hashtag;
and, ${\it freq}({\it hashtag}, {\it stance\_label})$ is the number of tweets that have that particular
hashtag and stance label.
We automatically extracted stance-indicative hashtags from the Stance Training set,
by considering only those hashtags that occurred at least five times and for which $H(hashtag) > 0.6$.
We will refer to this set of automatically compiled stance-indicative hashtags as {\it Automatic SI Hashtags}.
Table \ref{Tab:hashtag_example} lists examples.

We used both the Manual SI Hashtags and the Automatic\@ SI Hashtags as queries to select tweets from the Stance Domain Corpus.
(Recall that the Stance Domain Corpus is the large set of tweets pertaining to the five targets that was not manually labeled for stance.)
We will refer to the set of tweets in the domain corpus that have the Manual SI Hashtags as the {\it Manual Hashtag Corpus},
and those that have the Automatic\@ SI Hashtags as the {\it Automatic Hashtag Corpus}.
We then assign to each of these tweets the stance label associated with the stance-indicative hashtag they contain.
These noisy stance-labeled tweets can be used by a stance-classification system in two ways: (1) by including them as additional training data,
OR (2) by capturing words that are associated with a particular stance towards the target (word--stance associations) and 
words that are associated with a target (word--target associations), and using these associations to generate additional features
for classification.\footnote{Note that these word association features are akin to unigram features, except that they
are pre-extracted before applying the machine learning algorithm on the training corpus.}

On the one hand, method 1 seems promising because it lets the classifier directly use additional training data;
on the other hand, the additional training data is noisy and can have a
   very different class distribution than the manually labeled training and test sets. 
This means that the additional training data can impact the learned model disproportionately and adversely.
Thus we experiment with both methods.\footnote{We leave domain adaptation experiments
for future work.}

Table \ref{Tab:result_corpus} shows the results obtained on the Stance Test set when our stance classifier is
trained on various training sets. Observe that using additional training data
provides performance gains for three of the five targets. However,
marked improvements are observed only for `Hillary Clinton'.
It is possible, that in other test sets, the pseudo-labeled data is too noisy
to be incorporated as is. 
Thus, we next explore incorporating this pseudo-labeled data through additional features.

\begin{table*}
\tbl{F-scores of our supervised classifier (SVM with $n$-gram and target features) trained on different datasets. The highest scores for each column are shown in bold.\label{Tab:result_corpus}}{
\centering
 \resizebox{\textwidth}{!}{
\begin{tabular}{lrrrrrrrrrr}

\hline
 \bf Training Set       & Atheism	& Climate 	& Feminist  	& Hillary   	&Legal. of &F- &F-\\  
 			            & 			& Change 	& Movemt.  	& Clinton   	&Abortion &macroT &microT\\ \hline 
 
a. Stance Train Set    &       65.2 & \bf 42.2	& 57.7 	& 60.2 	& \bf 66.1 	& \bf 58.3	  & \bf 69.1 \\
 b. a. + Manual Hashtag Corpus     & 62.2 & \bf 42.2 & 50.5 & \bf 64.7 & 62.9 & 56.5 & 66.0\\
 c. a. + Automatic Hashtag Corpus   &\bf 65.8 & 40.2 & \bf 57.8 & 60.7 & 60.5 & 57.0 & 67.4\\
 \hline
\end{tabular}
}
}
\end{table*}

The association between a term and a particular stance towards the target is calculated
using pointwise mutual information (PMI) as shown below:\footnote{\citeN{turney2002thumbs} and \citeN{Kiritchenko2014} used similar measures for
word--sentiment associations.}
\vspace{-2mm}
\begin{equation}
 {\it PMI} (w, {\it stance\_label})=log_2 \frac{{\it freq}(w, {\it stance\_label}) *N}{{\it freq}(w) * {\it freq}({\it stance\_label})} 
\vspace{-2mm}
\end{equation}
\noindent where ${\it freq} (w, {\it stance\_label})$ is the number of times a term $w$ occurs in tweets that have
{\it stance\_label};
${\it freq} (w)$ is the frequency of $w$ in the corpus;
${\it freq}({\it stance\_label})$ is the number of tokens in tweets with label
$stance\_label$; and $N$ is the number of tokens in the corpus. 
When the system is trained on the Stance Training set, additional features are generated by taking the sum, min, and max
of the associations scores
for all the words in a tweet. 
Word--target association scores are calculated and used in a similar manner.

Table \ref{Tab:result_lex} shows the stance-classification results on the Stance Test set when 
using various word--association features extracted from the domain corpus.
Observe that the use of word--association features leads to improvements for all targets.
The improvements are particularly notable for `Atheism', `Feminist Movement', and `Legalization of Abortion'.
Also, the associations obtained from the Automatic\@ Hashtag Corpus are more informative to the classifier than those from the Manual Hashtag Corpus.

\begin{table}
 \tbl{F-scores for our classifiers that use word--associations extracted from the domain corpus. The highest scores in each column are shown in bold.\label{Tab:result_lex}}{
 \centering
\resizebox{\textwidth}{!}{
 \begin{tabular}{lrrrrrrrrrr}
 \hline
 \bf Features            & Atheism	& Climate 	& Feminist  	& Hillary   	&Legal. of &F- &F-\\ 
 			            & 			& Change 	& Movemt.  	& Clinton   	&Abortion &macroT &microT\\ \hline 
  a. $n$-grams + target 	& 65.2 &  42.2	& 57.7 	& 60.2 	& 66.1 	& 58.3	  &  69.1\\
    b. a. + associations (Manual Hashtags)  & & & & & \\
    $\;\;\;$ b1.    word--target associations    & 65.6 & 42.7 & 59.9 & 57.6 & 62.8 & 57.7 & 69.0 \\
    $\;\;\;$ b2.    word--stance associations    & 63.0 & 42.2 &58.3 &\bf 60.8 & 63.5 & 57.6 & 68.4\\
    $\;\;\;$ b3.    b1. + b2.                   & 65.9 & 42.7 & 59.0 & 56.9 & 64.0 & 57.7 & 68.7\\
    c. a. + associations (Automatic Hashtags)  & & & & & \\
    $\;\;\;$ c1.    word--target associations    & 64.5 &\bf 43.5 & 58.7 & 55.3 &\bf 68.8 & 58.1 &\bf 69.6\\
    $\;\;\;$ c2.    word--stance associations   & 65.1 & 42.4 & 59.1 & 59.8 & 64.3 & 58.1 & 69.2\\
    $\;\;\;$ c3.    c1. + c2.                 &  64.6 &\bf 43.5 & 58.8 & 56.7 & 67.5 & 58.2 & 69.5 \\
d. a. + associations (b1. + b2. + c1. + c2.)  &\bf 68.8 & 43.3 &\bf 60.8 & 56.2 & 64.1 &\bf 58.6 &\bf 69.6\\

 \hline
 \end{tabular}
 }
}
\end{table}

\subsection{Word Embeddings}

Word embeddings are low-dimensional real-valued vectors used to represent words in the vocabulary \cite{Bengio2001}. (The `low' dimensionality is relative to the vocabulary size, and using a few hundred dimensions is common.) 
A number of different language modeling techniques have been proposed to generate word embeddings, all of which require only a large corpus of text (e.g.,  \cite{collobert2008unified,mnih2009scalable}). Word embeddings have been successfully used as features in a number of tasks including sentiment analysis \cite{tang2014learning} and named entity recognition \cite{turian2010word}. Here we explore the use of large collections of tweets to generate word embeddings
as additional features for stance classification. We investigate whether they lead to further improvements over the results obtained by the best system configuration discussed in Section 6 --- SVM trained on the stance training set and using $n$-gram and target features.

We derive 100-dimensional word vectors using Word2Vec Skip-gram model \cite{mikolov2013distributed} trained over the Domain Corpus (the window size was set to 10, and the minimum count to 2). 
Given a training or test tweet, the word embedding features for the whole tweet are taken to be the component-wise averages of the word vectors for all the words appearing in the tweet.  

\begin{table*}[t!]
\tbl{Stance Classification: F-scores obtained by our classifier with additional word embedding features. The highest scores in each column are shown in bold.\label{tab:embed}}{
\centering
{\footnotesize
\begin{tabular}{lrrrrrrrrr}
\hline
 	&Atheism	& Climate 	& Feminist  	& Hillary   	&Legal. of	&F- &F-\\
\bf Classifier  	&    	& Change 	& Movemt. 	& Clinton	&Abortion  	&macroT &microT  \\ \hline
a. $n$-grams + target 	& 65.2 &  42.2	& 57.7 	&\bf 60.2 	& 66.1 	& 58.3	  &  69.1\\
b. a. + embeddings &\bf 68.3 &\bf 43.8 &\bf 58.4 & 57.8 &\bf 66.9 &\bf 59.0 &\bf 70.3\\
\hline
\end{tabular}
}
}
\label{Tab:result_features}
\end{table*}

Table \ref{tab:embed} shows stance classification results obtained using these word embedding features over and above the best configuration described in Section 6.
Observe that adding word embedding features improves results for all targets except `Hillary Clinton'.
Even though some teams participating in SemEval-2016 shared task on this dataset used word embeddings, 
their results are lower than those listed in Table \ref{tab:embed}. This is likely because they generated word embeddings from a generic corpus of tweets rather than tweets associated with the target (as is the case with the domain corpus). 

Overall, we observe that the three methods we tested here (adding noisy-labeled data as new training instances, adding noisy-labeled data through association features, or generating word embeddings) affect different subsets of data differently. 
For example, the `Hillary Clinton' subset of the test set benefited most from additional training data (Table \ref{Tab:result_corpus}) but failed to draw benefit from the embedding features. 
Such different behavior can be attributed to many possible reasons, such as the accuracy of hashtag-based labels, the class distribution of the new data, the size of the additional corpus, etc. 
Still, incorporating word embeddings seems a robust technique to improve the performance of stance detection in the presence of large unlabeled corpora. 

\section{Related Work}
\label{sec:relwork}
{\bf Stance Detection} Supervised learning has been used in almost all of the current approaches for stance classification, in which a large set of data has been collected and annotated in order to be used as training data for classifiers. In work by \citeN{somasundaran2010recognizing}, a lexicon for detecting argument trigger expressions was created and subsequently leveraged to identify arguments. These extracted arguments, together with sentiment expressions and their targets, were employed in a supervised learner as features for stance classification. 
\citeN{anand2011cats} deployed a rule-based classifier with several features such as unigrams, bigrams, punctuation marks, syntactic dependencies and the dialogic structure of the posts. The dialogic relations of agreements and disagreements between posts were exploited by \citeN{walker2012stance}. \citeN{faulkner2014automated} investigated the problem of detecting document-level stance in student essays by making use of two sets of features that are supposed to represent stance-taking language. 
\citeN{sobhaniargumentation} extracted arguments used in online news comments to detect stance.
 \citeN{djemili2014does} use a set of rules based on the syntax and discourse
 structure of the tweet to identify tweets that contain stance. 
\citeN{rajadesingan2014identifying} determine stance at user-level based on the
assumption that if several users retweet one pair of tweets about a
controversial topic, it is likely that they support the same side of a debate.

Existing datasets for stance detection were created from online debate forums like 4forums.com and createdebates.com \cite{somasundaran2010recognizing,walker2012corpus,hasan2013stance}. The majority of these debates are two-sided, and the data labels are often provided by the authors of the posts. Topics of these debates are mostly related to ideological controversial issues such as gay rights and abortion.

{\bf Sentiment Analysis and Opinion Mining} There is a vast amount of work in sentiment analysis of tweets, and we refer the reader to surveys \cite{PangL08,Liu12,mohammadsentiment}
and proceedings of recent shared task competitions \cite{SemEval2013task2,rosenthal2015semeval}.
Closely-related is the area of aspect based sentiment analysis (ABSA),
where the goal is to determine sentiment towards aspects of a product such as speed of processor and screen resolution of a cell phone.
We refer the reader to SemEval proceedings for related work on ABSA \cite{pontiki2015semeval,SemEval2014ABSA}.
\citeN{MohammadSemEval2013} and \citeN{Kiritchenko_SemEval2014} came first in the 2013 Sentiment in Twitter and 2014 SemEval ABSA shared tasks. We use most of the features they use
in our classifier. 

There has been considerable interest in analyzing
political tweets towards detecting sentiment, emotion, and purpose in
electoral tweets \cite{Mohammad2015political}, determining political
alignment of tweeters \cite{Golbeck11,Conover11b}, identifying
contentious issues and political opinions \cite{Maynard11}, detecting
the amount of polarization in the electorate \cite{Conover11}, and
even predicting the voting intentions or outcome of elections
\cite{Tumasjan10,Bermingham11,lampos2013user}. 

 There are other subtasks in opinion mining related to stance classification, such as
 biased language detection \cite{recasens2013linguistic,yano2010shedding}, perspective
 identification \cite{stance_1} and user classification based on their views
 \cite{kato2008taking}. Perspective identification was defined as the subjective
 evaluation of points of view \cite{stance_1}.  
\citeN{deng2014joint} suggested an
 unsupervised framework to detect implicit sentiment by inference over explicit sentiments
 and events that positively or negatively affect the theme. 
None of the prior work has created a dataset annotated for both stance and sentiment.
Neither has any work directly and substantially explored the relationship between stance and sentiment.

{\bf Textual Entailment} 
In textual entailment, the goal is to infer a textual statement (hypothesis) from a given source text \cite{dagan2004probabilistic}. Textual entailment is a core NLP building block, and has applications in question answering, machine translation, information retrieval and other tasks. It has received a lot of attention in the past decade, and we refer the reader to surveys \cite{androutsopoulos2010survey,dagan2013recognizing} and proceedings of recent challenges on recognizing textual entailment \cite{bentivogli2011seventh,marelli2014semeval,dzikovskajoint}. 

The task we explore in this paper, stance detection in tweets, can be viewed as another application of textual entailment, where the goal is to infer a person's opinion towards a given target based on a single tweet written by this person. In this special case of textual entailment, the hypotheses are always fixed (the person is either in favor of or against the target). Furthermore, we need to derive not only the meaning of the tweet, but also the attitude of the tweet's author.

{\bf Distant Supervision} Distant supervision makes use of indirectly labeled (or weakly labeled) data. This approach is widely used in automatic relation extraction, where information about pairs of related entities can be acquired from external knowledge sources such as Freebase or Wikipedia \cite{craven1999constructing,mintz2009distant}. Then, sentences containing both entities are considered positive examples for the corresponding relation. In sentiment and emotion analysis, weakly labeled data can be accumulated by using sentiment clues provided by the authors of the text--clues like emoticons and hashtags \cite{Go2009,Mohammad12}. Recently, distant supervision has been applied to topic classification \cite{husby2012topic,magdy2015bridging}, named entity recognition \cite{ritter2011named}, event extraction \cite{reschke2014event}, and semantic parsing \cite{parikh2015grounded}.

\section{Conclusions and Future Work}
\label{sec:conc}
We presented the first dataset of tweets annotated for both
stance towards given targets and for polarity of language.
The tweets are also annotated for whether opinion is expressed towards the given
target or towards another entity. 
Partitions of the stance-annotated data created as part of this project were used as training and test sets in a recent shared task competition on stance detection that received submissions from 19 teams. 
We proposed a simple, but effective stance detection system that obtained an F-score (70.3) higher than the one obtained by the more complex, best-performing system in the competition.  
We use a linear-kernel SVM classifier that leverages word and character $n$-grams as well as sentiment features drawn from available sentiment lexicons and word-embedding features drawn from additional unlabeled data. 

We presented a detailed analysis of the dataset
and conducted several experiments to tease out the interactions between 
stance and sentiment. 
Notably, we showed that sentiment features are not as effective for stance detection as they are for sentiment prediction. 
Moreover, an oracle system that had access to gold sentiment and target of opinion annotations was able to predict stance with an F-score of only 59.6\%.  
We also showed that even though 
humans are capable of detecting stance towards a given target from texts that
express opinion towards a different target, automatic systems perform poorly on such data.


In future work, we will explore the use of more sophisticated features 
(e.g., those derived from dependency parse trees and automatically
generated entity--entity relationship knowledge bases)
and more sophisticated classifiers (e.g., deep architectures that jointly model
stance, target of opinion, and sentiment).
We are interested in developing stance detection systems that do not require
stance-labeled instances for the target of interest, but instead, can learn from
existing stance-labeled instances for other targets in the same domain. 
We also want to model the ways in which stance is conveyed, and
how the distribution of stance towards a target changes over time.


 \begin{acks}
 \vspace*{-1mm}
 We thank Colin Cherry and Xiaodan Zhu for helpful discussions. The second author of this paper was supported by the Natural Sciences and Engineering Research Council of Canada under the CREATE program. 
\end{acks}
\vspace*{-2mm}

\bibliographystyle{ACM-Reference-Format-Journals}
\bibliography{references}
\vspace*{-5mm}
\received{January 2016}{..}{..}

\end{document}